\documentclass[conference]{IEEEtran}
\usepackage{graphicx}
\usepackage{amsmath}
\usepackage{amssymb}
\usepackage{hyperref}
\usepackage{cite}
\usepackage{booktabs}

\title{Arabic Large Language Models for Medical Text Generation}
\author{
    \IEEEauthorblockN{Abdulrahman Allam, Seif Ahmed, Ali Hamdi, Ammar Mohammed}
    \IEEEauthorblockA{
        \textit{October University for Modern Sciences \& Arts (MSA)} \\
        Giza, Egypt \\
        \{abdulrahman.atif, seifeldein.ahmed, ahamdi, ammohammed\}@msa.edu.eg
    }
}
\date{}

\begin{document}
\maketitle

\begin{abstract}
Efficient hospital management systems (HMS) are critical worldwide to address challenges such as overcrowding, limited resources, and poor availability of urgent health care. Existing methods often lack the ability to provide accurate, real-time medical advice, particularly for irregular inputs and underrepresented languages. To overcome these limitations, this study proposes an approach that fine-tunes large language models (LLMs) for Arabic medical text generation. The system is designed to assist patients by providing accurate medical advice, diagnoses, drug recommendations, and treatment plans based on user input. The research methodology required the collection of a unique dataset from social media platforms, capturing real-world medical conversations between patients and doctors. The dataset, which includes patient complaints together with medical advice, was properly cleaned and preprocessed to account for multiple Arabic dialects. Fine-tuning state-of-the-art generative models, such as Mistral-7B-Instruct-v0.2, LLaMA-2-7B, and GPT-2 Medium, optimized the system’s ability to generate reliable medical text. Results from evaluations indicate that the fine-tuned Mistral-7B model outperformed the other models, achieving average BERT (Bidirectional Encoder Representations from Transformers) Score values in precision, recall, and F1-scores of 68.5\%, 69.08\%, and 68.5\%, respectively. Comparative benchmarking and qualitative assessments validate the system’s ability to produce coherent and relevant medical replies to informal input. This study highlights the potential of generative artificial intelligence (AI) in advancing HMS, offering a scalable and adaptable solution for global healthcare challenges, especially in linguistically and culturally diverse environments.
\end{abstract}

\section{Introduction}
The increasing global demand for healthcare services, coupled with limited resources and rising patient expectations, highlights the critical need for advanced technologies that enable disease detection, resource allocation, and emergency reporting \cite{detmer1997computer,brown1984concept,rashid2025ai}. Current healthcare management systems (HMS) face significant challenges, including the inability to deliver fast and accurate medical support, effective resource distribution, and immediate reporting of critical cases. Training large language models (LLMs) for medical text creation provides an important opportunity for addressing these issues. By leveraging LLMs \cite{nazi2024large,peng2023study,salem2024enhancing,hamdi2024llm,hamdi2024riro}, healthcare systems can improve accuracy, scalability, and adaptability, bridging the gap between patients and efficient care delivery.

Despite progress in artificial intelligence (AI), existing HMS solutions generally rely on traditional machine learning (ML) models and rule-based systems, which often struggle to handle the complexity of medical data \cite{kuhl2022artificial,priyanka2024hospital}. These systems are constrained by the limited availability of high-quality, domain-specific medical datasets, especially for less-represented languages and contexts \cite{abdelhay2023deep,al2020nabiha,hosam2024lmrpa}. Moreover, traditional ML models have difficulty processing the unstructured, informal inputs commonly seen in patient feedback, resulting in reduced accuracy and reliability. The absence of robust solutions for integrating disease detection, resource allocation, and emergency reporting into a unified framework exacerbates these limitations \cite{namee2024enhancing,brown1984concept}.

To address these gaps, a dataset was curated that contains patient complaints and corresponding doctor responses sourced from social media platforms. These data represent authentic social interactions in which qualified doctors provide medical advice directly. Unlike conventional datasets, this collection reflects real-world language use, including unstructured, informal text and diverse linguistic variations \cite{denecke2009valuable,della2019health,abdelhay2023deep,abdellaif2024erpa}. Such data are essential for training LLMs capable of generating accurate, contextually relevant medical responses. However, the lack of fine-tuned LLMs specifically tailored for medical text generation remains a significant bottleneck in the adoption of generative AI for healthcare \cite{zhang2023generative,peng2023study}.

\begin{figure*}[h] 
    \centering
    \includegraphics[width=0.9\textwidth]{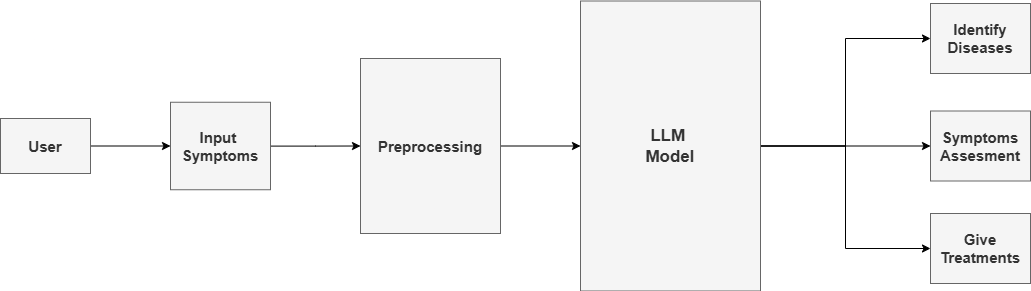}
    \caption{Explanation of the LLM-based system architecture.}
    \label{fig:ffff}
\end{figure*}

This study introduces a framework for fine-tuning LLMs in medical text creation, thereby facilitating more accurate disease detection, resource allocation, and emergency reporting. Figure~\ref{fig:ffff} illustrates the system architecture, where user inputs are preprocessed and then passed to the LLM, which produces outputs such as disease predictions, resource optimization strategies, and emergency alerts. This approach applies modern generative AI techniques to overcome the constraints of classical ML systems and address the complexity of unstructured medical data \cite{abdellaif2024lmrpa}.

The main contributions of this work are as follows:
\begin{enumerate}
    \item A curated dataset of Arabic medical conversations from social media platforms, capturing real-world patient complaints and doctor responses.
    \item A fine-tuned generative AI model for medical text generation, addressing the complexities of unstructured input and multiple Arabic dialects.
    \item A rigorous evaluation of the proposed model using metrics such as BERT Score (precision, recall, and F1-score), as well as qualitative evaluations of the generated medical responses.
\end{enumerate}

The remainder of this paper is organized as follows: Section~\ref{related_work} reviews related research on LLMs and HMS. Section~\ref{methodology} describes the dataset, model training procedures, and evaluation methods. Section~\ref{experimental} outlines the experimental design, and Section~\ref{discussion} discusses the results in comparison with baseline methods. Finally, Section~\ref{conclusion} provides concluding remarks and future research directions.

\section{Related Work}\label{related_work}

The use of large language models (LLMs) in healthcare has risen significantly due to their ability to enhance patient care while reducing the burden on healthcare infrastructure. These systems integrate natural language processing (NLP) and machine learning (ML) techniques to analyze user queries, provide medical suggestions, and optimize resource allocation \cite{nazi2024large, peng2023study, abdellatif2024lmv}. Despite these advancements, developing healthcare-focused LLMs in Arabic poses challenges resulting from linguistic diversity, limited data availability, and unstructured speech scenarios \cite{abdelhay2023deep, al2020nabiha}.

\subsection*{Challenges in Traditional Approaches}
Traditional healthcare chatbots have frequently relied on rule-based systems or classical ML approaches, which can perform well on structured, domain-specific tasks but often lack scalability and flexibility \cite{rekik2023, priyanka2024hospital, wassim2024llm}. For instance, a Tunisian medical chatbot showed high accuracy but was limited by a small dataset of curated cases and an inability to handle diverse dialects or informal queries \cite{rekik2023}. Similarly, chatbots in structured healthcare environments, such as PharmaGo \cite{gamage2021}, employed predictive models (e.g., LSTM networks) for structured data, making them less suitable for unstructured, real-world interactions \cite{priyanka2024hospital, anwar2025towards}.

\subsection*{Generative AI for Unstructured Medical Data}
Recent developments in generative artificial intelligence (AI) provide enhanced flexibility for handling unstructured inputs, making them promising tools for conversational healthcare settings. Generative models can produce diverse and contextually relevant responses, overcoming the limitations of rule-based systems and structured datasets \cite{zhang2023generative, nazi2024large, hamad2024asem}. However, many available generative AI solutions for healthcare are generic and not specifically tailored to Arabic medical contexts. Adapting these models to address Arabic’s linguistic complexities and varied dialects remains an underexplored area, requiring specialized datasets and comprehensive preprocessing methods \cite{al2020nabiha, abdelhay2023deep, salem2024enhancing}.

\subsection*{Study Contribution}
Building on previous research, this study presents a fine-tuned generative AI model developed specifically for Arabic medical chatbot applications. The proposed approach addresses challenges related to data scarcity, linguistic diversity, and unstructured user queries by compiling a real-world dataset of medical interactions and applying advanced generative models. This work demonstrates the potential of generative AI to improve healthcare services in historically underserved regions.

\section{Problem Formulation}
The lack of high-quality, domain-specific datasets—alongside difficulties in processing informal medical data—creates substantial hurdles for developing an effective Arabic medical chatbot. Unlike English, Arabic does not have the extensive resources needed to train models capable of handling multiple dialects and informal conversations \cite{al2020nabiha, abdelhay2023deep, anwar2025towards}. 

An additional issue involves the limited contextual detail in many user-generated medical queries. Numerous inquiries lack essential information, such as patient history or detailed symptom descriptions, making it challenging to generate precise medical responses. For instance, vague questions like \textit{“Why does my stomach hurt?”} often require more context to yield actionable recommendations \cite{salem2024enhancing, anwar2025towards}.

Traditional evaluation metrics, including precision, recall, and F1-score, may not fully capture the strengths of generative AI models. Measures based on exact matches between generated outputs and reference data can undervalue models that produce semantically accurate but textually distinct responses. Consequently, the evaluation strategy for this work integrates qualitative assessments of response coherence, logical accuracy, and contextual relevance.

Additionally, the unstructured nature of Arabic medical queries, coupled with linguistic variations, necessitates extensive preprocessing and model fine-tuning. Ensuring the accuracy of generated responses further requires validation mechanisms, such as the incorporation of verified medical sources or expert reviews, to mitigate the risk posed by incomplete or misleading user-generated inputs \cite{aldea2024using, zhang2023generative}. These challenges form the basis for designing and evaluating a scalable, domain-specific Arabic healthcare chatbot.

\section{Proposed Model}\label{methodology}

\subsection*{Dataset Collection and Description}
To leverage available data for training, a specialized dataset was constructed by compiling over 40,000 social media posts on medical topics. This collection features real-world medical questions, responses, and related metadata, reflecting informal, conversational language and dialectal variations \cite{denecke2009valuable, della2019health, abdelhay2023deep}. After preprocessing (detailed below), the dataset was refined to approximately 20,000 high-quality question--answer pairs suitable for model training.

A custom Python-based web scraping pipeline was used to gather the data, employing libraries such as Selenium for browser automation, BeautifulSoup for HTML parsing, and Pandas for data storage \cite{della2019health}. Facebook groups served as the primary source for extracting posts and replies. Adaptive scrolling, dynamic content handling, and pattern-based extraction techniques were applied to ensure dataset completeness \cite{abdelhay2023deep}.

\begin{figure}[h]
    \centering
    \includegraphics[width=0.5\textwidth]{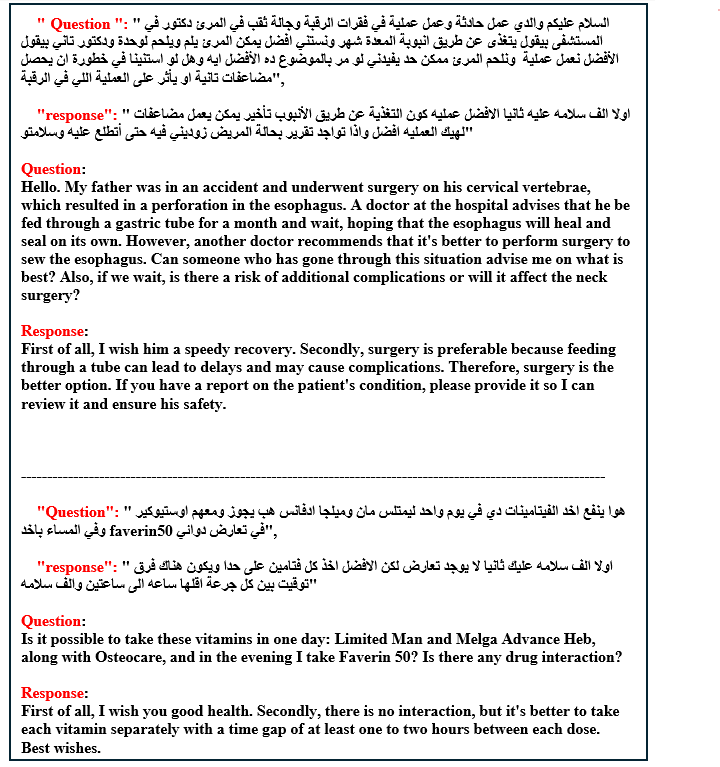}
    \caption{A sample from the curated dataset.}
    \label{fig:example_image}
\end{figure}

\subsection*{Preprocessing Pipeline}
The raw dataset underwent substantial preparation to improve quality and utility:
\begin{itemize}
    \item \textbf{Noise Removal:} Eliminated irrelevant content, including advertisements, emojis, and incomplete entries.
    \item \textbf{Normalization:} Standardized text to address dialectal differences, inconsistent spellings, and informal phrasing \cite{salem2024enhancing}.
    \item \textbf{Deduplication:} Removed redundant entries to improve data consistency.
\end{itemize}

Pattern recognition techniques and regular expressions were employed to correct typographical errors and repeated phrases \cite{abdelhay2023deep}. These steps produced a curated set of approximately 20,000 high-quality question--answer pairs, ensuring data relevance and structure \cite{anwar2025towards}.

\subsection*{Model Fine-Tuning}
Several pre-trained generative models were fine-tuned to adapt them to the Arabic medical domain:
\begin{itemize}
    \item \textbf{AraGPT2-Base:} A Generative Pretrained Transformer 2 (GPT-2)-based model specialized for Arabic text generation \cite{zhang2023generative}.
    \item \textbf{Meta-LLaMA/LLaMA-2-7B:} A high-performing transformer model offering scalability and adaptability across diverse tasks \cite{nazi2024large}.
    \item \textbf{BigScience/BLOOM-560M:} A multilingual model optimized for low-resource languages such as Arabic \cite{nova2023generative}.
    \item \textbf{Mistral-7B-Instruct-v0.2:} A generative model tailored for instruction-based tasks, demonstrating strong performance on unstructured queries \cite{peng2023study}.
    \item \textbf{OpenAI GPT-2 Medium:} A moderately sized generative model effective for general-purpose natural language generation \cite{kuhl2022artificial}.
\end{itemize}

The fine-tuning process made use of the Hugging Face Transformers library, incorporating low-rank adaptation (LoRA) to reduce computational costs while maintaining robust performance \cite{nova2023generative}. Each model was trained to recognize Arabic medical terminology, symptom labels, and user queries, enabling the generation of contextually relevant and diverse responses.

\subsection*{Evaluation}
The fine-tuned models were assessed using a combination of quantitative and qualitative methods:
\begin{itemize}
    \item \textbf{BERTScore:} Measured semantic similarity between generated responses and reference answers \cite{peng2023study}.
    \item \textbf{Qualitative Review:} Examined coherence, diversity, and contextual appropriateness of the generated text.
\end{itemize}

Overall, experimental results indicated that fine-tuned generative AI models performed effectively on informal and unstructured medical queries, surpassing traditional approaches in adaptability and robustness \cite{abdelhay2023deep, zhang2023generative}.

\section{Experimental Design}\label{experimental}

The experimental framework for this study was carefully constructed to apply state-of-the-art, pre-trained language models to the Arabic medical domain. By using advanced libraries, fine-tuning techniques, and dynamic training setups, the design aimed to optimize models for both performance and efficiency \cite{peng2023study}. Although traditional baseline models, such as rule-based systems or basic classification models, were not comprehensively tested due to time constraints, the evaluation approach was developed to thoroughly assess the performance of fine-tuned generative artificial intelligence (AI) models.

\subsection{Hugging Face Pretrained Models and Libraries}
This research utilized pre-trained models available on the Hugging Face platform \cite{nazi2024large}, in conjunction with the capabilities of the Hugging Face Transformers library for natural language processing (NLP) tasks \cite{abdelhay2023deep}. The Hugging Face AutoTokenizer library handled text tokenization, ensuring that the data was processed and fed into the models in an optimal format \cite{denecke2009valuable}.

\subsection{Fine-Tuning with Low-Rank Adaptation (LoRA)}
To enhance model performance while minimizing computational costs, Low-Rank Adaptation (LoRA) was employed during fine-tuning \cite{kuhl2022artificial}. LoRA allowed large language models to adapt to the specific characteristics of Arabic medical data without the overhead associated with full fine-tuning, making this approach particularly effective for domain-specific tasks \cite{zhang2023generative}.

\subsection{Baseline Models and Fine-Tuning Enhancements}
The performance of several baseline model variants was evaluated against their fine-tuned counterparts to determine the impact of domain-specific adaptation on Arabic medical chatbot tasks. Table \ref{tab:model_comparison} presents a comparison of base model and fine-tuned performance, highlighting noteworthy improvements achieved through fine-tuning across all models.

\begin{table}[h!]
\centering
\caption{Comparison of Base Model and Fine-Tuned Model Performance}
\begin{tabular}{lcc}
\toprule
\textbf{Model Name} & \textbf{Base Model (\%)} & \textbf{Fine-Tuned (\%)} \\ 
\midrule
Meta-LLaMA/LLaMA-2-7B       & 64.00 & 66.50 \\ 
AraGPT2-Base (148M)         & 55.45 & 63.00 \\ 
OpenAI GPT-2 Medium (355M)  & 53.61 & 64.00 \\ 
Mistral-7B-Instruct-v0.2    & 65.77 & 68.50 \\ 
BigScience/BLOOM-560M       & 50.12 & 64.03 \\ 
\bottomrule
\end{tabular}
\vspace{3mm}
\label{tab:model_comparison}
\vspace{-3mm}
\end{table}

\paragraph{Meta-LLaMA/LLaMA-2-7B and Mistral-7B-Instruct-v0.2}  
These models achieved the highest baseline performance (64.00 for LLaMA and 65.77 for Mistral), likely due to large parameter sizes and robust pre-training. While they demonstrated strong general comprehension of Arabic text and context, specificity in medical contexts was limited. After fine-tuning, their performance reached 66.50 and 68.50, respectively, making them the top-performing models in this comparison.

\paragraph{AraGPT2-Base (148M)}  
Although AraGPT2 displayed lower baseline performance (55.45) compared to LLaMA and Mistral, it consistently produced grammatically correct Arabic text. Its responses, however, lacked adequate medical context. Fine-tuning boosted its performance to 63.00, aligning more closely with domain-specific needs while preserving fluency in Arabic.

\paragraph{OpenAI GPT-2 Medium (355M)}  
The GPT-2 model scored 53.61 at baseline, showing moderate generative capabilities but limited proficiency with medical terminology. Fine-tuning elevated its performance to 64.00, reflecting its adaptability to specialized tasks.

\paragraph{BigScience/Bloom-560M}  
Bloom registered the lowest baseline score (50.12), likely stemming from its multilingual focus and limited specialization in Arabic. Fine-tuning significantly improved its performance to 64.03, emphasizing the importance of domain-specific adaptation for this model.

\subsection{Training Process}
Each model---including \texttt{aubmindlab/aragpt2-base}, \texttt{meta-llama/Llama-2-7b-hf}, \texttt{bigscience/bloom-560m}, \texttt{Mistral-7B-Instruct-v0.2}, and \texttt{OpenAI GPT-2 Medium (355M)}---was fine-tuned using the curated Arabic medical dataset. This process aligned the outputs of each model with the unique requirements of medical contexts, ensuring the generation of accurate, relevant responses. Key training considerations included:
\begin{itemize}
    \item \textbf{Learning Rate Schedule:} A cosine schedule with warmup steps supported stable convergence.
    \item \textbf{Batch Size:} Balanced computational efficiency with resource constraints.
    \item \textbf{Epochs:} Selected to avoid overfitting while enabling sufficient domain learning.
\end{itemize}

Generative AI models were chosen due to their capacity to handle unstructured and informal queries more effectively than rule-based systems or classifiers. Rather than relying on fixed outputs, generative approaches produce contextually varied responses, which is crucial for medical dialogues and better suits real-world applications in Arabic healthcare scenarios.

\subsection{Training Configuration}
The training pipeline was managed through Hugging Face’s \texttt{TrainingArguments} class \cite{nova2023generative}. Primary parameters included:

\paragraph{Output Directory}  
Results were saved under \texttt{./results} to organize model checkpoints.

\paragraph{Evaluation and Save Strategy}  
Evaluation and checkpoint saving were performed at the end of each epoch to preserve the best-performing model.

\paragraph{Batch Sizes}  
Both training and evaluation batch sizes were set to 8, striking a balance between computational efficiency and resource limitations.

\paragraph{Warmup Steps}  
A warmup of 200 steps provided smoother convergence in the early stages of training \cite{salem2024enhancing}.

\paragraph{Mixed Precision}  
Mixed-precision training (FP16) was employed to increase speed and reduce memory usage without sacrificing performance \cite{khan2024medimatch}.

\section{Results}\label{discussion} 
The results of the Arabic medical large language model (LLM) demonstrate promising performance in understanding and generating responses within the Arabic medical domain. Evaluation was conducted using Average BERT Precision, BERT Recall, and F\textsubscript{1}-Score \cite{salem2024enhancing}, ensuring accuracy, reliability, and contextual appropriateness \cite{abdelhay2023deep}.

\subsection{Model Performance}
Five models were evaluated on subsets of the dataset, primarily due to computational constraints for large models that generate free-form responses \cite{peng2023study}. Key results appear in Table~\ref{tab:model_results}.

\begin{table}[h!]
\centering
\caption{Performance Metrics of Different LLM Models Based on Average BERT Score}
\label{tab:model_results}
\resizebox{\linewidth}{!}{%
\begin{tabular}{lccc}
\toprule
\textbf{Model Name} & \multicolumn{3}{c}{\textbf{Average BERT Score}} \\ 
\cmidrule(lr){2-4}
 & \textbf{Precision (\%)} & \textbf{Recall (\%)} & \textbf{F1 Score (\%)} \\ 
\midrule
Meta-LLaMA/LLaMA-2-7B      & 66.50  & 69.00   & 67.25 \\ 
AraGPT2-Base (148M)        & 63.00  & 67.02  & 65.04 \\ 
OpenAI GPT-2 Medium (355M) & 64.00  & 67.00   & 65.07 \\ 
Mistral-7B-Instruct-v0.2   & 68.50  & 69.08  & 68.50 \\ 
BigScience/BLOOM-560M      & 64.03 & 67.29  & 65.60 \\ 
\bottomrule
\end{tabular}
}
\vspace{2mm}
\end{table}

A central component of the evaluation was the BERTScore metric, which leverages contextual embeddings to measure the semantic similarity between generated and reference responses \cite{salem2024enhancing}. This metric offers a nuanced view of the model’s capacity to produce contextually relevant and semantically coherent output.

\subsection{BERTScore and Equation Definitions}
\begin{itemize}
    \item \textbf{BERTScore:} Calculates precision, recall, and F\textsubscript{1}-Score by comparing token embeddings of the model’s output with reference tokens.
    
    \item \textbf{Precision:} Assesses how closely the candidate tokens align with reference tokens in a semantic space \cite{salem2024enhancing}.
    \begin{equation}\label{eq:precision}
        \text{Precision}_{\text{BERTScore}} 
        = \frac{1}{\mathit{N_c}} \sum_{i=1}^{\mathit{N_c}} \max_{j} S_{i,j}
    \end{equation}
    In \eqref{eq:precision}, \(\mathit{N_c}\) is the total number of tokens in the generated candidate, and \(S_{i,j}\) is the semantic similarity between the \(i\)-th candidate token and the \(j\)-th reference token.

    \item \textbf{Recall:} Evaluates how comprehensively the reference tokens are covered by the generated response \cite{salem2024enhancing}.
    \begin{equation}\label{eq:recall}
        \text{Recall}_{\text{BERTScore}} 
        = \frac{1}{\mathit{N_r}} \sum_{j=1}^{\mathit{N_r}} \max_{i} S_{i,j}
    \end{equation}
    In \eqref{eq:recall}, \(\mathit{N_r}\) is the total number of reference tokens, and \(S_{i,j}\) is defined as in \eqref{eq:precision}, measuring semantic similarity between tokens.

    \item \textbf{F\textsubscript{1}-Score:} Combines precision and recall into a single metric that accounts for both relevance and completeness \cite{salem2024enhancing}.
    \begin{equation}\label{eq:f1}
        \text{F}_1^{\text{BERTScore}} 
        = 2 \cdot \frac{\text{Precision}_{\text{BERTScore}} \cdot \text{Recall}_{\text{BERTScore}}}{\text{Precision}_{\text{BERTScore}} + \text{Recall}_{\text{BERTScore}}}
    \end{equation}
\end{itemize}

Throughout these equations, \(\text{Precision}_{\text{BERTScore}}\), \(\text{Recall}_{\text{BERTScore}}\), and \(\text{F}_1^{\text{BERTScore}}\) refer to the BERTScore-based precision, recall, and F\textsubscript{1}-Score, respectively.

\subsection{Observations}
\textbf{1) Best-Performing Models:}  
Mistral-7B-Instruct-v0.2 and Meta-LLaMA/LLaMA-2-7B demonstrated the highest F\textsubscript{1}-Scores, indicating strong fluency in generating grammatically correct Arabic text and handling common medical queries \cite{abdelhay2023deep}. These models appear well-suited for queries featuring straightforward symptom descriptions and matching dataset entries.

\begin{figure}[h]
    \centering
    \includegraphics[width=0.5\textwidth]{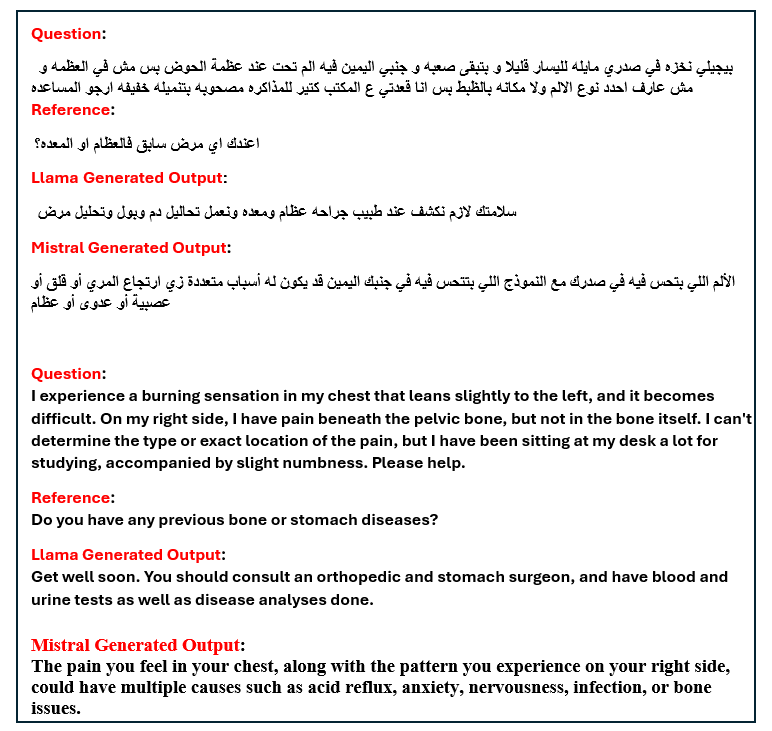}
    \caption{Comparison of Responses from LLaMA-2-7B and Mistral-7B Models}
    \label{fig:example_image}
\end{figure}

\begin{itemize}
    \item \textbf{Strengths:} Mistral-7B-Instruct-v0.2 and Meta-LLaMA/LLaMA-2-7B-hf generated contextually relevant, grammatically precise responses \cite{zhang2023generative, salem2024enhancing}. Smaller models such as AraGPT2-Base performed respectably in resource-constrained environments \cite{priyanka2024hospital, salem2024enhancing}.  
    \item \textbf{Limitations:} Handling ambiguous or highly specialized medical queries remained challenging for all models \cite{denecke2009valuable}. Reliance on user-generated data of variable quality also restricted output reliability \cite{priyanka2024hospital, abdelhay2023deep}. Enhanced data curation and enrichment could mitigate this issue.
\end{itemize}

\textbf{2) Efficiency vs. Accuracy Trade-Off:}  
AraGPT2-Base (148M) achieved an F\textsubscript{1}-Score of 65.04\%, highlighting a reasonable balance between model size and performance \cite{salem2024enhancing}. BigScience/BLOOM-560M demonstrated slightly higher metrics (65.60\%), showing how domain-specific fine-tuning can substantially boost multilingual models \cite{peng2023study}.

\subsection{Implications}
The top F\textsubscript{1}-Scores (e.g., 68.50\% from Mistral-7B-Instruct-v0.2) underline the potential of large-scale language models to address Arabic healthcare queries \cite{salem2024enhancing, priyanka2024hospital}. Although additional refinements—such as verifying medical data and improving handling of complex or ambiguous queries—are necessary, these findings provide a strong foundation for future exploration \cite{zhang2023generative}. The demonstrated performance of these models supports their practicality for real-world healthcare applications, particularly in regions where AI-based systems can enhance resource allocation and accessibility \cite{anwar2025towards}.

\section{Conclusion}\label{conclusion}
Development of an Arabic medical LLM represents a crucial advance in enhancing healthcare accessibility for Arabic-speaking communities. This study presented a methodical approach to addressing challenges that include limited, high-quality Arabic medical data and user input containing multiple dialects. By curating a real-world social media dataset, applying extensive preprocessing, and fine-tuning state-of-the-art generative models, a scalable solution was tailored to the Arabic medical domain.
Models such as Mistral-7B-Instruct-v0.2, LLaMA-2-7B, BLOOM-560M, AraGPT2-Base, and OpenAI GPT-2 Medium were employed, with Low-Rank Adaptation (LoRA) used to optimize training efficiency. Evaluations based on precision, recall, F\textsubscript{1}-Score, and qualitative assessments revealed each model’s capacity to address both informal and domain-specific queries effectively. 
These findings underscore the transformative potential of generative AI in bridging healthcare gaps across Arabic-speaking regions. Beyond the technical achievements, this work highlights the need for continuous improvements in data reliability, evaluation strategies, and real-world validation. Addressing these areas can further elevate model performance, bolstering trust and adoption in clinical and non-clinical healthcare settings.

\section{Acknowledgment}
Heartfelt gratitude is extended to AiTech AU, \textit{AiTech for Artificial Intelligence and Software Development} (\url{https://aitech.net.au}), for funding this research, providing technical support, and enabling its successful completion.

\bibliographystyle{IEEEtran}

\begin{thebibliography}{10}
\providecommand{\url}[1]{#1}
\csname url@samestyle\endcsname
\providecommand{\newblock}{\relax}
\providecommand{\bibinfo}[2]{#2}
\providecommand{\BIBentrySTDinterwordspacing}{\spaceskip=0pt\relax}
\providecommand{\BIBentryALTinterwordstretchfactor}{4}
\providecommand{\BIBentryALTinterwordspacing}{\spaceskip=\fontdimen2\font plus
\BIBentryALTinterwordstretchfactor\fontdimen3\font minus \fontdimen4\font\relax}
\providecommand{\BIBforeignlanguage}[2]{{%
\expandafter\ifx\csname l@#1\endcsname\relax
\typeout{** WARNING: IEEEtran.bst: No hyphenation pattern has been}%
\typeout{** loaded for the language `#1'. Using the pattern for}%
\typeout{** the default language instead.}%
\else
\language=\csname l@#1\endcsname
\fi
#2}}
\providecommand{\BIBdecl}{\relax}
\BIBdecl

\bibitem{detmer1997computer}
D.~E. Detmer, E.~B. Steen, and R.~S. Dick, ``The computer-based patient record: an essential technology for health care,'' 1997.

\bibitem{brown1984concept}
T.~C. Brown, ``The concept of value in resource allocation,'' \emph{Land economics}, vol.~60, no.~3, pp. 231--246, 1984.

\bibitem{rashid2025ai}
M.~Rashid and M.~Sharma, ``Ai-assisted diagnosis and treatment planning—a discussion of how ai can assist healthcare professionals in making more accurate diagnoses and treatment plans for diseases,'' \emph{AI in Disease Detection: Advancements and Applications}, pp. 313--336, 2025.

\bibitem{nazi2024large}
Z.~A. Nazi and W.~Peng, ``Large language models in healthcare and medical domain: A review,'' in \emph{Informatics}, vol.~11, no.~3.\hskip 1em plus 0.5em minus 0.4em\relax MDPI, 2024, p.~57.

\bibitem{peng2023study}
C.~Peng, X.~Yang, A.~Chen, K.~E. Smith, N.~PourNejatian, A.~B. Costa, C.~Martin, M.~G. Flores, Y.~Zhang, T.~Magoc \emph{et~al.}, ``A study of generative large language model for medical research and healthcare,'' \emph{NPJ digital medicine}, vol.~6, no.~1, p. 210, 2023.

\bibitem{salem2024enhancing}
L.~A. Salem, T.~Shishtawy, N.~El-Attar \emph{et~al.}, ``Enhancing healthcare management: A case study of a medical chatbot in egypt,'' \emph{Benha Journal of Applied Sciences}, vol.~9, no.~5, pp. 199--210, 2024.

\bibitem{hamdi2024llm}
A.~Hamdi, A.~A. Mazrou, and M.~Shaltout, ``Llm-sem: A sentiment-based student engagement metric using llms for e-learning platforms,'' \emph{arXiv preprint arXiv:2412.13765}, 2024.

\bibitem{hamdi2024riro}
A.~Hamdi, H.~Kassab, M.~Bahaa, and M.~Mohamed, ``Riro: Reshaping inputs, refining outputs unlocking the potential of large language models in data-scarce contexts,'' \emph{arXiv preprint arXiv:2412.15254}, 2024.

\bibitem{kuhl2022artificial}
N.~K{\"u}hl, M.~Schemmer, M.~Goutier, and G.~Satzger, ``Artificial intelligence and machine learning,'' \emph{Electronic Markets}, vol.~32, no.~4, pp. 2235--2244, 2022.

\bibitem{priyanka2024hospital}
M.~L. PRIYANKA, B.~T.~R. NAYAKA, B.~S. RAKSHITHA, B.~VAMSHI, and B.~M.~K. YADAV, ``Hospital management system with chatbot,'' \emph{International Journal of Mechanical Engineering Research and Technology}, vol.~16, no.~2, pp. 144--154, 2024.

\bibitem{abdelhay2023deep}
M.~Abdelhay, A.~Mohammed, and H.~A. Hefny, ``Deep learning for arabic healthcare: Medicalbot,'' \emph{Social Network Analysis and Mining}, vol.~13, no.~1, p.~71, 2023.

\bibitem{al2020nabiha}
D.~Al-Ghadhban and N.~Al-Twairesh, ``Nabiha: an arabic dialect chatbot,'' \emph{International Journal of Advanced Computer Science and Applications}, vol.~11, no.~3, 2020.

\bibitem{hosam2024lmrpa}
O.~Hosam~Abdellaif, A.~Nader, and A.~Hamdi, ``Lmrpa: Large language model-driven efficient robotic process automation for ocr,'' \emph{arXiv e-prints}, pp. arXiv--2412, 2024.

\bibitem{namee2024enhancing}
K.~Namee, T.~Prempranee, W.~Phonngam, R.~Kaewsaeng-On, and A.~Meny, ``Enhancing hospital bed management through chatbots: Integrating dialogflow and firebase cloud messaging for real-time bed availability and reservation system,'' in \emph{2024 22nd International Conference on ICT and Knowledge Engineering (ICT\&KE)}.\hskip 1em plus 0.5em minus 0.4em\relax IEEE, 2024, pp. 1--6.

\bibitem{denecke2009valuable}
K.~Denecke and W.~Nejdl, ``How valuable is medical social media data? content analysis of the medical web,'' \emph{Information Sciences}, vol. 179, no.~12, pp. 1870--1880, 2009.

\bibitem{della2019health}
S.~Della~Rosa, F.~Sen \emph{et~al.}, ``Health topics on facebook groups: content analysis of posts in multiple sclerosis communities,'' \emph{Interactive Journal of Medical Research}, vol.~8, no.~1, p. e10146, 2019.

\bibitem{abdellaif2024erpa}
O.~H. Abdellaif, A.~N. Hassan, and A.~Hamdi, ``Erpa: Efficient rpa model integrating ocr and llms for intelligent document processing,'' in \emph{2024 International Mobile, Intelligent, and Ubiquitous Computing Conference (MIUCC)}.\hskip 1em plus 0.5em minus 0.4em\relax IEEE, 2024, pp. 295--300.

\bibitem{zhang2023generative}
P.~Zhang and M.~N. Kamel~Boulos, ``Generative ai in medicine and healthcare: promises, opportunities and challenges,'' \emph{Future Internet}, vol.~15, no.~9, p. 286, 2023.

\bibitem{abdellaif2024lmrpa}
O.~H. Abdellaif, A.~Nader, and A.~Hamdi, ``Lmrpa: Large language model-driven efficient robotic process automation for ocr,'' \emph{arXiv preprint arXiv:2412.18063}, 2024.

\bibitem{abdellatif2024lmv}
O.~Abdellatif, A.~Ayman, and A.~Hamdi, ``Lmv-rpa: Large model voting-based robotic process automation,'' \emph{arXiv preprint arXiv:2412.17965}, 2024.

\bibitem{rekik2023}
A.~Rekik, M.~Chebbi, and N.~Boughanmi, ``A medical chatbot for tunisian dialect using a rule-based and machine learning approach,'' \emph{Journal of Medical Informatics}, vol.~12, no.~3, pp. 45--56, 2023.

\bibitem{wassim2024llm}
L.~Wassim, K.~Mohamed, and A.~Hamdi, ``Llm-daas: Llm-driven drone-as-a-service operations from text user requests,'' \emph{arXiv preprint arXiv:2412.11672}, 2024.

\bibitem{gamage2021}
R.~G. Gamage, K.~U. Senadeera, N.~S. Bandara, K.~Y. Abeywardena, D.~D. Diyamullage, and N.~Amarasena, ``Pharmago: An online pharmaceutical ordering platform,'' in \emph{2021 3rd International Conference on Advancements in Computing (ICAC)}, 2021, pp. 364--370.

\bibitem{anwar2025towards}
J.~Anwar, P.~Nadi, and N.~Seddik, ``Towards building a chatbot-based first aid service in arabic language,'' \emph{Journal of Advanced Research in Applied Sciences and Engineering Technology}, vol.~45, no.~2, pp. 1--10, 2025.

\bibitem{hamad2024asem}
O.~Hamad, K.~Shaban, and A.~Hamdi, ``Asem: Enhancing empathy in chatbot through attention-based sentiment and emotion modeling,'' in \emph{Proceedings of the 2024 Joint International Conference on Computational Linguistics, Language Resources and Evaluation (LREC-COLING 2024)}, 2024, pp. 1588--1601.

\bibitem{aldea2024using}
M.~Aldea, P.~Rolland, S.~Simon, A.~Poplu, M.~Wartelle, B.~Vignal, J.-C. Louis, F.~Lion, A.~Borie, D.~Planchard \emph{et~al.}, ``Using ai to automatically process data from unstructured health records of patients with lung cancer,'' \emph{Cancer Research}, vol.~84, no. 6\_Supplement, pp. 3569--3569, 2024.

\bibitem{nova2023generative}
K.~Nova, ``Generative ai in healthcare: advancements in electronic health records, facilitating medical languages, and personalized patient care,'' \emph{Journal of Advanced Analytics in Healthcare Management}, vol.~7, no.~1, pp. 115--131, 2023.

\bibitem{khan2024medimatch}
S.~Khan, A.~Saify, S.~Gosaliya, D.~Jain, and M.~J. Zalte, ``Medimatch: Ai-driven drug recommendation system,'' in \emph{2024 2nd International Conference on Sustainable Computing and Smart Systems (ICSCSS)}.\hskip 1em plus 0.5em minus 0.4em\relax IEEE, 2024, pp. 1342--1349.

\end{thebibliography}

\end{document}